\documentclass{article} % For LaTeX2e
\usepackage{gpl_iclr2022_conference,times}

\usepackage{amsmath,amsfonts,bm}

\def\eqref#1{equation~\ref{#1}}
\def\1{\bm{1}}

\DeclareMathAlphabet{\mathsfit}{\encodingdefault}{\sfdefault}{m}{sl}
\SetMathAlphabet{\mathsfit}{bold}{\encodingdefault}{\sfdefault}{bx}{n}

\usepackage{hyperref}
\usepackage{url}
\usepackage[dvips]{graphicx}
\usepackage{subfig}
\usepackage{algorithm}
\usepackage{algpseudocode}
\usepackage{bbm}

\usepackage{natbib}

\begin{document}
\title{Safer Autonomous Driving in a Stochastic, Partially-Observable Environment by Hierarchical Contingency Planning}

% Authors must not appear in the submitted version. They should be hidden
% as long as the \iclrfinalcopy macro remains commented out below.
% Non-anonymous submissions will be rejected without review.

\author{Ugo Lecerf \thanks{correspondance to: \texttt{ugo.lecerf@renault.fr}} \\ Renault Software Labs \\ EURECOM \And Christelle Yemdji Tchassi \\ Renault Software Labs \And Pietro Michiardi \\ EURECOM}

\newcommand{\fix}{\marginpar{FIX}}
\newcommand{\new}{\marginpar{NEW}}

\iclrfinalcopy % Uncomment for camera-ready version, but NOT for submission.

\maketitle

\begin{abstract}

When learning to act in a stochastic, partially observable environment, an intelligent agent should be prepared to anticipate a change in its belief of the environment state, and be capable of adapting its actions on-the-fly to changing conditions. As humans, we are able to form contingency plans when learning a task with the explicit aim of being able to correct errors in the initial control, and hence prove useful if ever there is a sudden change in our perception of the environment which requires immediate corrective action.

This is especially the case for autonomous vehicles (AVs) navigating real-world situations where safety is paramount, and a strong ability to react to a changing belief about the environment is truly needed. 

In this paper we explore an end-to-end approach, from training to execution, for learning robust contingency plans and combining them with a hierarchical planner to obtain a robust agent policy in an autonomous navigation task where other vehicles' behaviours are unknown, and the agent's belief about these behaviours is subject to sudden, last-second change. We show that our approach results in robust, safe behaviour in a partially observable, stochastic environment, generalizing well over environment dynamics not seen during training.
\end{abstract}

\section{Introduction}

Developing a controller for AVs to be used in a real-life navigation environment poses a number of challenges including perception, and modeling environment dynamics \citep{end_to_end_UC_quantification_Michelmore, AD_Survey_common_practices2020}. The limitations in vehicle sensors, as well as the stochastic nature of inter-vehicle interactions, introduce a level of uncertainty in autonomous navigation tasks which hinders the ability to control an agent from the standpoint of both safety, and adequate performance. Sensor information, as well as other drivers' intentions, are subject to sudden change and a robust navigation algorithm must be able to safely adapt on-the-fly to these unforeseen changes.

Stochastic environments can be modeled as Partially-Observable Markov Decision Processes (POMDPs) \citep{Kochenderfer_book}. Solving POMDPs is challenging, yet possible, for example using methods combining learning and planning such as \citet{brechtel14, Hoel2019}. Having access to a model of the environment dynamics allows us to use planning algorithms, such as tree-search methods \citep{mcts_survey}, alongside learning to both increase sample efficiency, and have access to a better representation of the environment's state-space structure \citep{Machado_LaplacianOptionDiscovry}. In cases where planning is possible it is much easier for an agent to predict the outcome of its actions and hence better adapt to eventual changes in the state-space \citep{mcallister_pilco_for_pomdps}. However, a change in the values of the stochastic model parameters (e.g. a change in a vehicle’s sensor accuracy, unplanned scene obstruction, or simply an unforeseen behaviour from another vehicle) may induce a sharp drop in the agent’s performance due to its inability to generalize well to new environment parameters.

An approach for tackling this issue has been to design controllers to have a high capacity for generalization: instead of attempting to learn over the entire space of possible environments, we design an algorithm to act well enough over the whole space, having trained only on a tractable subset of environment configurations. One such example is known as contingency planning, whereby contingency plans are put in place to specifically counter stochastic environments in which failing to prepare for possible problems in advance can be an expensive mistake \citep{Pryor1996PlanningForContingencies}.

\textbf{Our contribution.}
In this work we make the following contributions:
\begin{itemize}
    \item We introduce a method for learning a contingency policy concurrently to the optimal policy during training, such that the former is well-adapted to serve as a contingency plan.
    \item We combine learned policies with a high-level model based controller, and experimentally show through an autonomous navigation task that our approach is able to achieve a much safer agent performance in the case of a stochastic environment, while sacrificing a minimal amount of performance.
\end{itemize}

\section{Background}

Deep reinforcement learning (RL) is a method for learning control algorithms in a weakly-supervised manner (by means of a reward signal). Implementing an RL algorithm requires us to model the environment as a Markov Decision Process (MDP), defined by the following elements: A finite set of states $s\in\mathcal{S}$, indexed by the timestep at which they are encountered: $s_t$. A finite set of actions $a\in \mathcal{A}$, also indexed by their respective timesteps: $a_t$. A transition model $\mathcal{T}(s,a,s'):\mathcal{S}\times\mathcal{A}\times\mathcal{S} \rightarrow [0,1]$, representing the probability of passing from $s$ to $s'$ after taking action $a_t$, $P(s_{t+1}=s'|s_t=s, a_t=a)$. An immediate reward function $R(s,a,s'):\mathcal{S} \times \mathcal{A}\times\mathcal{S} \rightarrow \mathbb{R}$ and discount factor $\gamma\in (0,1]$, controlling the weight in value of states further along the Markov chain. A POMDP is further augmented by an observation model $\mathcal{O}$, when we no longer have access to the true state $s_t$, but to an observation thereof $o_t\sim\mathcal{O}(s_t)$.

Actions in the POMDP are taken by a policy $\pi :\mathcal{O} \rightarrow \mathcal{A}$ mapping observations to actions. The value of an observation under a policy $\pi$, is given by the value function $V^\pi:\mathcal{O}\rightarrow \mathbb{R}$, which represents the expected future discounted sum of rewards, if policy $\pi$ is followed from the true state $s$, of which we have an observation $o\sim O(s)$:
\begin{gather}
\label{eq:value_fn_definition}
V^\pi(o):=\mathbb{E}\left[\sum^\infty_{t=0}\gamma^t R(s_t,a_t,s_{t+1}) \right], \\ 
s_0=s, o_t\sim O(s_t), a_t=\pi(o_t), s_{t+1}\sim\mathcal{T}(s_t,a_t). \nonumber
\end{gather}
Actions are chosen by the policy $\pi$, so as to maximize the action-value function $Q^\pi:\mathcal{O}\times\mathcal{A} \rightarrow \mathbb{R}$ which assigns values to actions according to the value of the states that are reached under $\pi$:
\begin{gather}
\label{eq:q_fn}
    Q^\pi(o_t,a_t):=\mathbb{E}_{s_{t+1}}\left[R(s_t,a_t,s_{t+1})+\gamma\cdot V^\pi(o_{t+1})\right], \\ 
    s_{t+1}\sim \mathcal{T}(s_t,a_t), o_{t+1}\sim O(s_{t+1}). \nonumber
\end{gather}

We use $Q^\pi$ in order to define the optimal policy, which we denote $\pi^*$, as the policy taking actions that maximizes $Q^\pi$: $\pi^*(o):=\text{arg}\max_{a\in\mathcal{A}}Q^{\pi^*}(o,a)$. Equation (\ref{eq:q_fn}) highlights that the action-value function is a sort of one-step look-ahead of the value of the next possible state $s_{t+1}$, in order to determine the value of actions in the current observation $o_t$. We denote the sequence of observations (trajectory through observation-space) visited by a policy $\pi$ during an episode as: 
\begin{gather}
\tau_\pi:=\left\{o_t\right\}_{t\in[0,T]}, \\
s_0\in\mathcal{S}, o_t\sim O(s_t), a_t=\pi(o_t), s_{t+1}\sim\mathcal{T}(s_t,a_t). \nonumber
\end{gather}

\subsection{Training an Off-Policy Deep Q Network}
The off-policy $Q$-learning algorithm \citep{mnih2015} is well-suited to learning a policy's $Q$-function in the discrete action space which we consider. Training off-policy allows us to train on past transitions stored in a replay buffer \citep{schaul2015prioritized} which both stabilizes training, and increases sample efficiency since a single transition sample may be used during multiple training steps.  We perform a stochastic gradient descent on the MSE between the $Q$-estimate for the current step, and the discounted 1-step `lookahead' $Q$-estimate summed with the transition reward:

$$\mathcal{L}(o_t, a_t, r_t, o_{t+1})=\left(Q_\theta(o_t,a_t) - \left[r_t + \gamma Q_{\theta'}(o_{t+1}, \text{arg}\max_{a'} Q_{\theta}(o_{t+1},a')\right]\right)^2,$$

where $\theta$ are the current parameters for the $Q$-network, and $\theta'$ are parameters that are `frozen' for a certain amount of steps. This is known as double $Q$-learning using target networks \citep{hasselt15_double} and reduces the variance of gradient updates.

\section{Related Work}
\textbf{Hierarchical RL.}
Hierarchical reinforcement learning  \citep{FeudalRL_Hinton,sutton99options} is a promising approach for helping RL algorithms generalize better in increasingly complex environments. Several works apply the hierarchical structure of control to navigation tasks, which lend themselves well to modular controllers \citep{Fisac2018_game_theory_hirarchical_RLforAVs}. Approaches such as \citet{Andreas_Policy_Sketches, data_efficient_HRL_NACHUM} use sub-task or goal labelling in order to explicitly learn policies that are able to generalize through goal-space. Our approach differs where we don't wish to learn different goals, rather striving to attain the same goal under modified environment conditions, moreover, we aim to learn alternative strategies without having to subtask or goal labels. Works such as \citet{Machado_LaplacianOptionDiscovry, zhang2021hierarchical} also have the same aim of self-discovering strategies to be used in a hierarchical controller, though our method differs by targeting the value function of the agent, through additional reward terms. Similar to our approach, \citet{Cunningham_Multipol_decision_making_in_UC} uses a form of voting through policy simulation, though our work also integrates training the contingency policy with robustness to environment modifications in mind.

\textbf{Contingency planning.}
Some approaches use an estimation of the confidence of the actions proposed by an agent's policy \citep{Bouton19_scene_decomposition, clements2019estimating, hoel20_IV_ensemble_uncertainty} to determine whether or not an agent's policy is sufficiently good in the current environment state. When this is not the case, control is typically given to a separate, often open-loop controller looking to mitigate any possible negative behaviour if failure cannot be avoided otherwise \citep{dalal2018, Angelos_out_of_distribution2020}. One issue with open-loop contingency plans -- or any open-loop policy in general -- is that they do not take into account the closed-loop nature of most real-world environments, whose dynamics are dependent on the actions of the agent and may themselves fail if not implemented carefully. \citet{contingencies_from_observations, 2021_Killing_robust_bi_lane} tackle the problem of high environment uncertainty by prioritizing information gathering if the agent is too uncertain about its policy's outcome. These approaches choose to approach by default with caution, if ever there is missing or uncertain information in the agent's input space. \citet{kumar2020one_solution} seeks to learn a set of policies which are collectively robust to changes in environment dynamics, through the use of latent-conditioned policies \citep{Eysenbach_diversity_is_all_you_need}. Our approach is similar, though we use an explicit metric on trajectories to ensure diversity in contingency behaviour rather than a learnt descriminator function. Furthurmore we train our contingency plan to perform well when environment parameters change during the execution phase, which couples well with an on-line high-level controller.

\section{Learning Contingency Policies}
\textbf{Problem statement.} 
We wish to learn, on one hand, $\pi^*$, the optimal policy in our given environment, and on the other, $\pi_1$, a contingency policy able to navigate more safely through the environment, at the cost of performance, if ever there is high uncertainty linked to following the optimal policy.

\subsection{Contingency Policies}
\label{sec:contingency_plans}
Given the nature of a contingency plan, it must aim to exploit trajectories through different areas of state-space than trajectories from the optimal policy. Since we consider cases where typically local uncertainties in the navigation task disrupt the performance of the optimal policy, in order for the contingency plan to stay viable it should have trajectories through sufficiently different areas of state-space to avoid areas of high uncertainty. For example a contingency plan to high-speed trajectories, will often be more conservative and favor low-speed policies.

\textbf{Using reward augmentation.}
Our approach for learning $\pi^*$ and $\pi_1$ concurrently, is to augment the reward function by a term which we will refer to as a \emph{trajectory penalty}, $R^{pen}$, which is proportional to the similarity in expected observation-space trajectories between two policies. The aim of augmenting the reward in this way is for the policy of the contingency agent to be incentivized to solve the environment through different areas of state-space than the optimal policy. With this aim in mind we define a metric on agents' observation trajectories:
\begin{equation}
    \mathcal{M}(\tau_{\pi_1}, \mathbb{E}\left[\tau_{\pi^*}\right]) := \int \Big \vert \nu\left(\phi\left(\tau_{\pi_1}\right)\right) - \nu\left(\phi\left(\mathbb{E}\left[\tau_{\pi^*}\right]\right)\right) 
    \Big \vert d\phi(o), 
    \label{eq:metric}
\end{equation}
where $\phi$ is an observation-feature function, and $\nu$ is the density function over observation features. It is possible to remain the most general possible with $\phi(o)=o$, however with some domain knowledge we can modify $\phi$ to retain only the features which best define the `distinctness' of an agent's trajectory.

Designing the penalty term $R^{pen}$ to be inversely proportional to the trajectory metric with respect to a reference agent, we can define it as:
\begin{equation}
R^{pen}_{\pi^*}(\tau_{\pi_1}) := -\frac{\alpha}{\mathcal{M}(\tau_{\pi_1},\mathbb{E}\left[\tau_{\pi^*}\right]) + \delta}, 
\label{eq:negative_rwd_implementation}
\end{equation}

where $\alpha$ and $\delta$ determine the relative weight of the trajectory penalty, with respect to the regular reward function $R$. A lower value for $\alpha$ will hardly penalize the contingency policy for having a similar observation distribution to the reference agent, whereas higher weighting will make it seek a highly different trajectory, disregarding the original objective of the task given by the regular reward function. The value function for $\pi_1$ becomes:

$$V^{\pi_1}(o):=\underset{\tau\sim\pi_1}{\mathbb{E}}\left[R(\tau)+R^{pen}_{\pi^*}(\tau)\right], \quad \tau\vert_{t=0}=o,$$
where $\tau\sim\pi_1$ is the expected observation-space trajectory under $\pi_1$. We denote $\Pi$, the set of all policies (optimal and contingency) learned in the POMDP. Stochastic parameters for the POMDP are stored in a vector $\boldsymbol{\mu}$. Appendix \ref{app:alg1} presents the pseudo-code for our proposed algorithm.

\textbf{Contingency policy's domain.}
When building a hierarchical control framework, much of the generalization capabilities come from the correct identification of the environment parameters, and then executing the relevant policy \citep{HRL_2021_survey}).
When a new objective is introduced to the agent or environment parameters change causing new environment dynamics, making the optimal policy unable to perform well, we must ensure that the contingency policy is able to compensate for the initial actions taken by the optimal policy. When the observation-space gets more complex, we may no longer assume that the learned contingency policy has sufficient knowledge over its domain such that it is able to correctly control the agent from any such `hand-off' state.

This is linked to the exploration trade-off made during training, where we explicitly limit an agent's exploration of the available state-space in the interest of making learning tractable \citep{SuttonF1998PracticalRL}). Due to limited exploration by design, this limits the domain on which a policy's $Q$-function is well-learnt. This isn't an issue for navigating from the starting state to each objective, however there will be some areas of state-space, less sampled by the agent, where the $Q$-function will have a greater error. This may be equivalently seen as a lack of exploration and insufficient training of the RL agent, however in more complex environments time and computing constraints prevent us from training the $Q$-function over the policy's entire domain.

To encourage good contingency policy performance in so-called `hand-off' states, our approach is to modify the initial state distribution for the contingency agent. We achieve this by sampling states from the optimal agent's replay buffer, and adding them to the contingency agent's initial state distribution. Given the initial state distribution $p(s_0)$, we define the new initial state distribution for the contingency agent as: $p^{\pi_1}(s_0)=(1-\beta)p(s_0) + \beta p(\tilde{s}_{\pi^*})$, where $p(\tilde{s}_{\pi^*})$ is the uniform distribution over states in $\pi^*$'s replay buffer, and $\beta\in [0,1]$ is a parameter controlling the ratio of initial states sampled from the regular task intialization to ones sampled from the optimal's replay buffer.

Moreover, our aim with this replay buffer intialization is to help the robustness of the contingency agent over the most probable `hand-off' points with the optimal agent when the latter is unable to handle the current environment parameters. To this end, we may further use domain knowledge to constrain the states sampled from $p(\tilde{s}_{\pi})$ such that they best represent these possible `hand-off' points. Appendix \ref{app:replay_buffer_init} includes more details about the replay buffer initialization.

\subsection{Hierarchical Controller}

\begin{figure}[h!]
    \centering
    \includegraphics[width=0.8\linewidth]{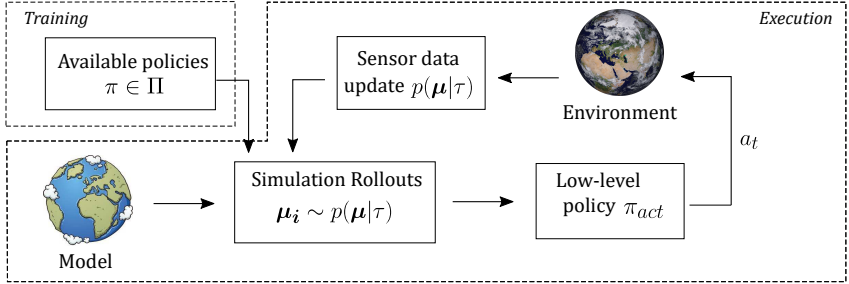}
    \caption{Structure of hierarchical controller composed of available policies and model-based planner (High-level policy selection).}
    \label{fig:hi_fn_diagram}
\end{figure}

Contingency plans alone are not sufficient: a higher-level controller is required, capable of selecting the best policy with respect to the current environment observation. In our approach, we combine the learned optimal and contingency policies with a model-based planner, in order to increase the robustness and safety of the acting agent with respect to environment uncertainty. Figure \ref{fig:hi_fn_diagram} shows the structure of the hierarchical controller, whose high-level policy selection is in fact the planner module responsible for estimating the safety of each of the available policies $\pi\in\Pi$, and selecting the one with the lowest estimated chance of failure. A pseudo-code description for the hierarchical controller is provided in appendix \ref{app:alg2}.

\subsection{Planner}
The planner's role is to estimate the chance of failure for each $\pi\in\Pi$ available to the controller. The approach we retain for our purposes is straightforward: we perform roll-outs over the set of possible environment parameters $\boldsymbol{\mu}$ for each $\pi$, and return an estimated failure rate based on those roll-outs which will define which policy is selected by the planner.
This process estimates the probability of policy failure given the current belief about environment stochastic parameters $\boldsymbol{\mu}$. Let $P(s_{fail}|\pi,\boldsymbol{\mu})$ denote the probability that policy $\pi$ will fail for a given $\boldsymbol{\mu}$. The planner's goal is to select a policy according to: 
$$
\pi= \text{arg}\min_{\pi\in\Pi} P(s_{fail}|\pi,\boldsymbol{\mu}^*),    
$$
where $\boldsymbol{\mu}^*$ are the true (unknown) parameter values. We may approximate it by sampling from a probability density function conditioned on the history of agent observations over the course of the episode $\boldsymbol{\mu}_i\sim p(\boldsymbol{\mu}|\tau)$. More details on sampling $\boldsymbol{\mu}_i$'s are given in appendix \ref{app:sampling_mu}. Let $C(\pi,\boldsymbol{\mu}_i)\in[0,1]$ represent whether or not there is a failure (collision) from policy $\pi$, after roll-out. Then:
$$
P(s_{fail}|\pi,\boldsymbol{\mu}^*)\approx \frac{1}{M}\sum_{i=1}^M C(\pi,\boldsymbol{\mu}_i),
$$
for $M$ samples of $\boldsymbol{\mu}_i$. We note that the quality of the approximation relies on how well the $\boldsymbol{\mu}_i$'s are sampled. The closer they are to $\boldsymbol{\mu}^*$, the better the planner will be able to estimate the true probability of failure for each policy. We use a simple approach consisting in eliminating $\boldsymbol{\mu}_i$'s from the sampling pool, if targets are observed behaving in contradiction to the considered environment parameters (based on vehicle speed, in our navigation task).

\section{Experiments}
\subsection{Simulation Environment}
\begin{figure}[tpb]
    \centering
    \subfloat[\centering $t=3$s]{{\includegraphics[width=2.5cm]{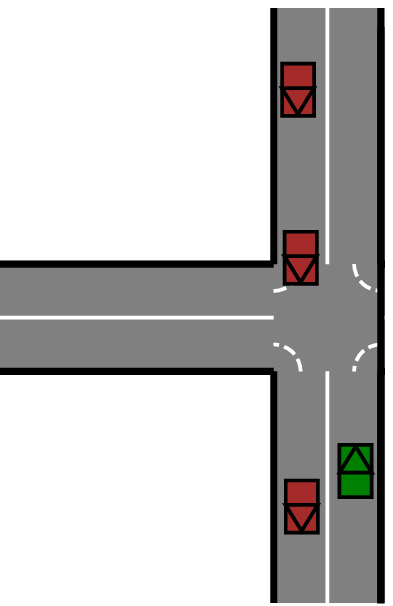} }}%
    \hspace{3cm}
    \subfloat[\centering $t=5$s]{{\includegraphics[width=2.5cm]{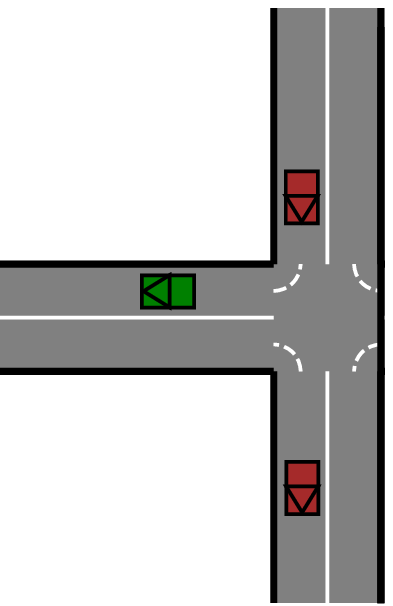} }}%
    \caption{Navigation task: ego makes a left turn across the intersection with oncoming traffic. Target vehicles may either be aggressive (i.e. disregarding presence of ego in intersection) or cooperative (i.e. slowing down if ego is close to intersection point).}%
    \label{fig:mdp_cross_capture}%
\end{figure}

We evaluate our approach in a common autonomous navigation task \citep{brechtel14,Hubann_driving_UC_environment, Bouton19_scene_decomposition, Risk_sensitive_Bernhard_2019_IV,contingencies_from_observations}). Figure \ref{fig:mdp_cross_capture} shows two frames of the environment with oncoming vehicles in the intersection. In navigation tasks, the controllable agent is usually referred to as the ego whereas the other vehicles are referred to as targets. In this task the ego must adjust its speed in order to pass through the intersection without colliding with any of the oncoming targets. The optimal policy with respect to the reward function (\ref{eq:reward_function}) is to pass through the intersection as fast as possible while avoiding collisions. We expect a contingency policy to be useful in this scenario, when there is a sudden change in estimated target behaviour by the planner and the ego will have to either slow down to let an aggressive target through, or speed up if it seems the target is slowing down too much which may also cause a collision.

\textbf{Ego and target behaviours.}
The ego's action space is a set of longitudinal acceleration values: $a\in \mathcal{A}=\{-4,-2,-1,0,1,2\}$ m/s$^{2}$. Target vehicles may have either a cooperative, or aggressive  style of driving, this behaviour being unobservable by the ego. An aggressive target will speed through the intersection disregarding the presence of the ego vehicle, whereas a cooperative target will slow down if ever the ego vehicle approaches the intersection, although it will not come to a complete stop if ever ego halts in the middle of the road.

\textbf{Reward function.}
To penalize collisions and encourage faster episode termination, the step-reward $r_t$ is set-up as follows per time step $t$:
\begin{equation}
r_t=
\left\{\begin{matrix}
-5 & \text{if collision} \\
-0.1 & \text{otherwise}
\end{matrix}\right. .
\label{eq:reward_function}
\end{equation}

\textbf{Stochastic environment parameters.}
The environment dynamics depend on the targets' behaviours. The behaviour for each target vehicle is a random variable $B_i$ representing degree of aggressiveness, which are collected in $\boldsymbol{\mu}$ where $\boldsymbol{\mu}|_i=B_i$ for $i\in[1,N]$ for $N$ targets. In practice we use either $b_i=0$ for a cooperative target, or $b_i=1$ for an aggressive target.

\subsection{Implementation Details}
\textbf{Target behaviours.}
During training we set $b_i=1, \enspace \forall i\in[1,N]$. Both optimal and contingency policies are trained on these environment parameters.

\textbf{Trajectory penalty.}
In practice we replace the expectation operator $\mathbb{E}\left[\tau_\pi\right]$ in (\ref{eq:metric}) by the mean value over the last 100 samples of $\pi^*$'s replay buffer. In this environment we use the speed of the ego vehicle as an observation feature $\phi(o)=\dot{x}_{ego}$. Trajectory penalty scaling factors used are: $\alpha = 3, \delta=0.1 \text{  ,}$ which are fixed by a rough initial sweep.

\textbf{Adjusting the return values for different initial states.}
We use a modified initialization for the contingency agent to increase its capabilities for compensating erroneous behaviour from a previous policy. We initialize 50\% of the contingency agent's episodes in this way, during the entire training phase. Forcing the contingency policy to start the episode in states sampled from the optimal policy's replay buffer will incur greater trajectory penalties due to the higher trajectory proximity. However, we do not need to to add an explicit compensation term: the contingency agent is still learning its true action-value function $Q^{\pi_1}$. We take into consideration that the trajectory penalty term $R^{pen}$ will reduce the mean computed performance score, so it is important to look at the training score and trajectory penalties separately when comparing performance from both policies. To compensate for varying episode length due to a different initial state, the contingency policy is attributed the cumulated rewards corresponding to the sampled initial state from the replay buffer.

\section{Results}
\begin{figure}[h!]
    \centering
    \subfloat[\centering]{{\includegraphics[width=7cm]{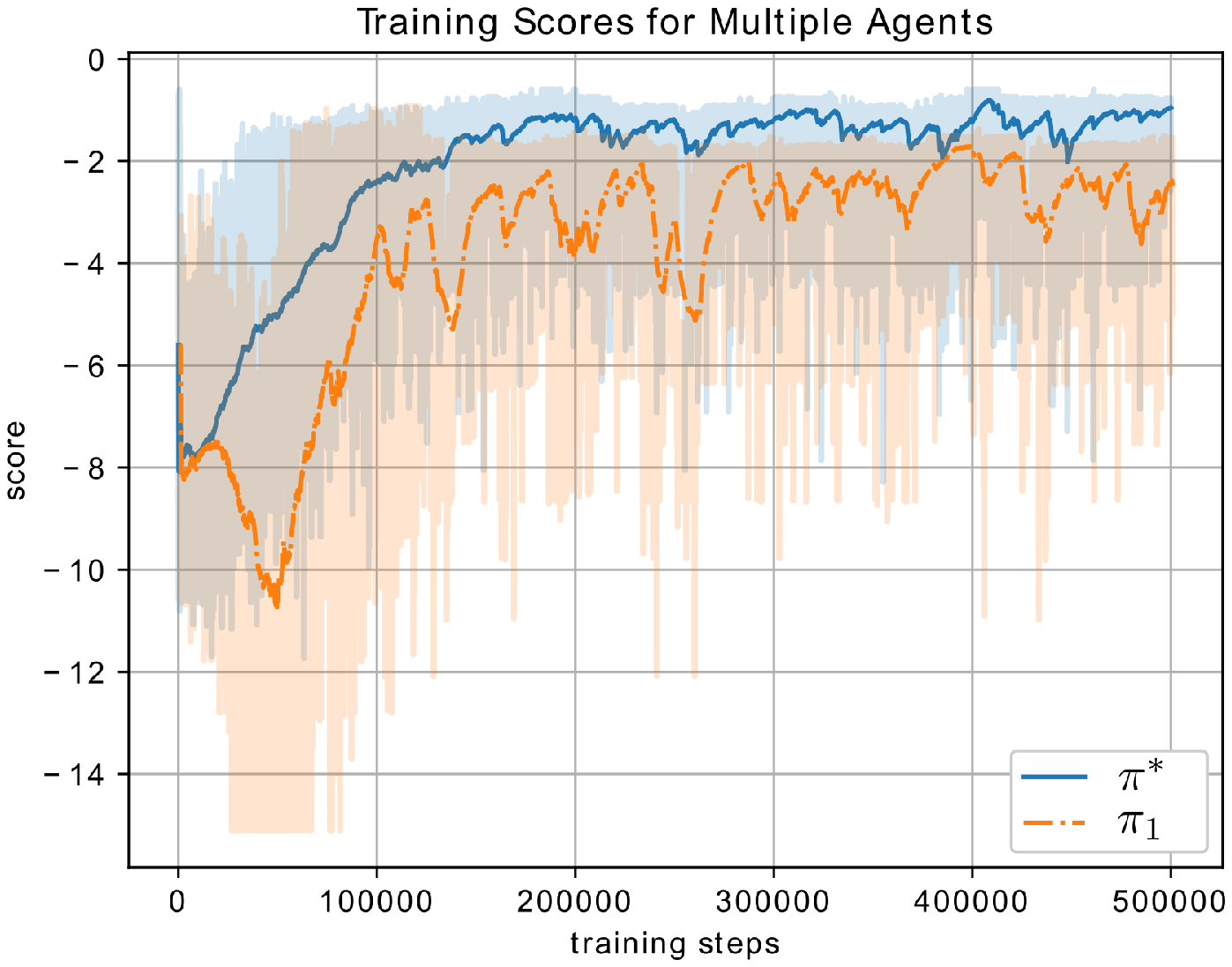} }}%
    \subfloat[\centering]{{\includegraphics[width=7cm]{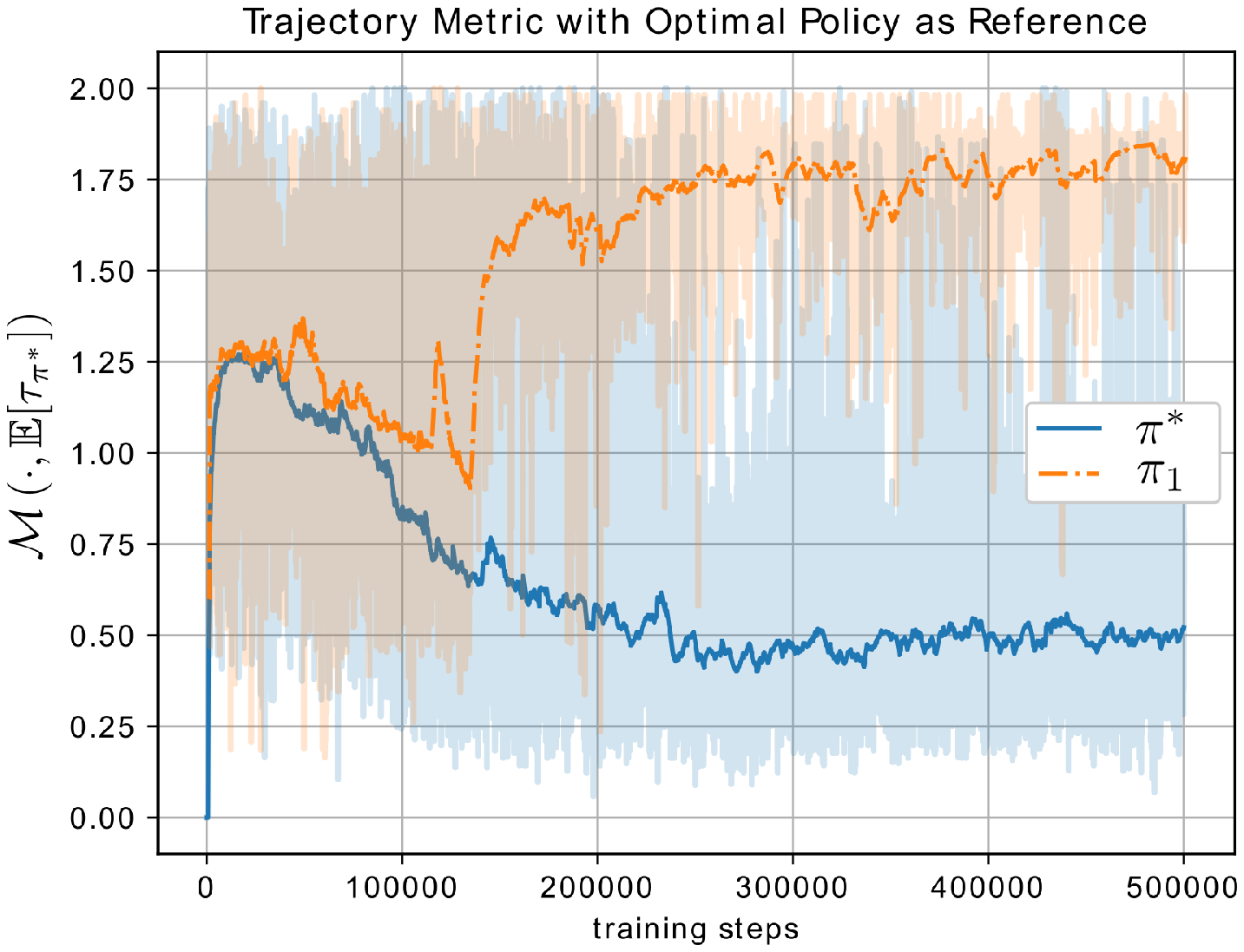} }}%
    \caption{(a) Training scores for both agents during training phase. This figure does not take into account the trajectory penalties $R^{pen}$ for $\pi_1$, only the regular rewards $R$. (b) Evolution of computed trajectory metric term $\mathcal{M}(\cdot, \mathbb{E}\left[\tau_{\pi^*}\right])$ for optimal and contingency policies. Although computed for both policies, the resulting trajectory penalty is only attributed to $\pi_1$.}%
    \label{fig:tr_results}%
\end{figure}

\textbf{Training a contingency policy with replay buffer.} 
Figure \ref{fig:tr_results}a shows the scores achieved by both the optimal and contingency policies during training. We can see $\pi^*$ learning until reaching its optimal performance score after approximately 200k training steps. $\pi_1$'s optimization objective depends on an additional $R^{pen}$ term, which is not stable until the observation-density in the replay buffer of $\pi^*$ becomes approximately static.  This is why we see the initial training curve of $\pi_1$ lag behind that of $\pi^*$. Once $\pi^*$'s expected observation trajectory, $\mathbb{E}\left[\tau_{\pi^*}\right]$, has a low enough variance then $\pi_1$ can make some meaningful progress towards its contingency objective. Looking at $\pi_1$'s training curve in our case this appears to happen around the 50k step mark. Even once $\pi_1$ has reached its top performance, it continues to be more jittery than $\pi^*$. This is due to the sampled replay buffer initializations.

\textbf{Trajectory metric during training.}
Figure \ref{fig:tr_results}b shows the evolution of the trajectory metric (\ref{eq:metric}). The values for $\pi^*$ are shown for comparison, but not actually included in its reward during training. The shape for $\pi^*$ is to be expected, as the latest trajectory will initially be different from the mean trajectory sampled from its own replay buffer. However as its parameters $\theta$ converge and exploration is weaned away, the episodic trajectory will get closer to the mean trajectory sampled from the replay buffer.  

We notice that initially both policies have similar trajectories $\tau$. This is due to the high degree of initial random exploration. There is a sharp rise in the metric for $\pi_1$ around the 150k mark, and we may deduce that at this point the most recent trajectories $\tau_{\pi_1}$ are sufficiently different to the observations present in $\pi^*$'s replay buffer such that the metric between the two increases. We may conclude that this is where the contingency policy is converging to a contingency plan that has a different trajectory from that of the optimal policy (notably lower-speed trajectories). 

\textbf{Hierarchical planner safety performance.}
To evaluate the overall performance of our hierarchical controller, we compute the number of successes vs. failures on the navigation task, over a range of environment parameter values $\boldsymbol{\mu}$ for different agents. Table \ref{tab:safety_results} compiles the results for each tested controller, over an environment parameter sample size of $M=200$.

\begin{table}[h!]
\caption{Controller performances for a range of agents. Success rate gives ratio of number of successes vs. failures over all sampled environment configurations. Average score gives us the mean of scores obtained in cases where the agent does not fail. Results are obtained by averaging 4 runs over the same random seed.}

\label{tab:safety_results}
\begin{center}
\begin{tabular}{l|l|l}
\textbf{Controllers} & \textbf{Success rate} & \textbf{Average Score} \\
\hline & & \\
$\mathbf{\pi^*}$ & 0.496 & -1.200 \\
\hline
$\mathbf{\pi_1}$ without replay buffer init. & 0.885 & -1.500 \\
\hline
$\mathbf{\pi_1}$ with replay buffer init. & 0.954 & -1.466 \\
\hline
H-control without replay buffer init. & 0.890 & -1.380 \\
\hline
H-control with replay buffer init. & $\boldsymbol{1.000}$ & $\boldsymbol{-1.238}$ \\

\end{tabular}
\end{center}
\end{table}

From our results we clearly deduce that a single agent has a hard time generalizing to new target behaviours, even though it may have achieved optimal performance within its training environment. Although $\pi^*$ has the highest average score in cases when it does not fail, it does not generalize well and performs very poorly in unseen environment instances. In our navigation task, the contingency policy that is learnt corresponds to the ego having a more conservative driving attitude; this explains why the success rate is higher for $\pi_1$ with respect to $\pi^*$, although the average score decreases due to sacrificing performance for safer behaviour. When adding the replay buffer initialization to $\pi_1$, the success rate further increases due to the increased ability of the contingency agent to generalize to greater areas of observation-space.

Compared to the individual policies, we expect the hierarchical controller to perform better, due to its access to a model-based planner combined with both policies. Interestingly, we find that the contingency policy with replay buffer initialization has a higher success rate than the hierarchical controller using $\pi_1$ without the initialization. This highlights the importance of generalization when designing safe, robust algorithms. Finally, we see that the highest success rate is achieved by our proposed approach of combining optimal, and robust contingency policies. More importantly, though success rates are similar with the single $\pi_1$ with replay buffer initialization, we are able to obtain a good performance score close to the optimal policy, due to $\pi^*$ being available to the hierarchical controller. This demonstrates how our approach is able to ensure much safer behaviour in unseen environment configurations than a single policy, without sacrificing performance by being overly cautious.

\section{Conclusion}
In conclusion, we have presented an approach for learning multiple policies in an autonomous navigation task and adapting the approach to specifically learn a robust contingency policy, which when combined with a model-based planner, is able to increase the robustness of the agent with respect to stochastic environment parameters. In our intersection use-case, we are able to reach a rate of no collisions for any (sampled) environment configuration, even when we only had access to a single one of these configurations during training. 
We acknowledge that the planner module has a simple behavioural prediction for target vehicles, and although it is sufficient in our simulation environment to obtain good performance, better detection of the environment's stochastic parameters $\boldsymbol{\mu}$ will increase the robustness of the overall agent, and ultimately the effectiveness of having an available contingency policy.

One main advantage of this hybrid approach is the ability to separate performance and safety in an RL framework. Whereas using a single reward function and relying on reward engineering to obtain the correct behaviour can be arbitrary, in this case we are able to optimize for performance in a complex environment, and ensure a high level safety in unseen instances of environment dynamics without having to tweak performance and safety terms in a single reward function.

\newpage

\bibliographystyle{plainnat}
\bibliography{references}

\appendix
\section{Implementation Details}
\subsection{Training Algorithm}
\label{app:alg1}
Algorithm \ref{alg:pol_training} gives an overview of the training regime used with the reward penalty, in order to learn contingency policies alongside the optimal. Notably, we attribute $r^{pen}$ to the final sample at the end of the episode played by the contingency agent. $\mathbbm{1}_{[\pi==\pi_1]}$ stands for the indicator function, indicating whether the current training agent is the contingency agent $\pi_1$, or not. 

Both agents train concurrently, alternating training episodes. However the contingency agent $\pi_1$ starts its training only when the replay buffer of $\pi^*$ is full. This is reflected in the training curves, though it is negligible with respect to the order of magnitude of total training samples. Computing the value for $=R^{pen}_{\pi^*}\left(\tau_{\pi_1}\right)$ before $\pi^*$'s expected trajectory memory buffer has converged to a stable value means that the MDP being solved by $\pi_1$ is initially greatly changing, though we did not investigate the effects of different training scheduling for the contingency agent in this work.

In the initial state distribution $p(s_0)$ during training, the ego always starts at the same position, whereas the initial positions of the target vehicles are randomized.

\begin{algorithm}
\caption{Training Contingency Policies}
\label{alg:pol_training}
\begin{algorithmic}[1]
\State Init $\Pi=\{\pi^*, \pi_1\}$
\While{not converged}
    \For{$\pi\in\Pi$}
        \State $s_0\sim p^\pi(s_0)$  \Comment{init. episode state}
        \State $o_0\sim \mathcal{O}(s_0)$
        \State $\tau_\pi = \{o_0\}$
        \While{episode not terminated} \Comment{play episode}
            \State $a_t=\pi(o_t)$
            \State $s_{t+1}\sim\mathcal{T}(s_t,a_t)$ \Comment{environment step}
            \State $r_t=R(s_t,a_t,s_{t+1})$  \Comment{step reward}
            \State $o_{t+1} \sim O(s_{t+1})$ 
            \State $\tau_\pi=\tau_\pi \cup \{o_{t+1}\}$
        \EndWhile
        \State $r^{pen}=R^{pen}_{\pi^*}\left(\tau\right)$ \Comment{compute reward penalty according to (\ref{eq:negative_rwd_implementation})}
        \State $r_T = r_T + \mathbbm{1}_{[\pi==\pi_1]} r^{pen}$ \Comment{ attribute $r^{pen}$ only to contingency agent}
        \For{$t\in [0,T]$}
            \State Memory$\left(\pi\right)\gets \left(o_t,a_t,r_t,o_{t+1}\right)$ \Comment{store samples in agent replay buffer}
        \EndFor
    \EndFor
\EndWhile
\end{algorithmic}
\end{algorithm}

\subsection{Replay Buffer Initialization}
\label{app:replay_buffer_init}
We mention in section \ref{sec:contingency_plans} the need for a contingency plan to function well at potential `hand-off' points, if the high-level controller switches to the contingency plan late in the episode. To increase the likelihood that $p(\tilde{s}_{\pi^*})$ well represents these states, using domain knowledge, we constrain it to have a uniform density only over states where the ego vehicle has not yet passed the intersection. This avoids the contingency policy learning to act once the ego has passed the intersection which is not useful in our use-case:
$$p(\tilde{s}_{\pi^*})=\mathcal{U}(\tilde{\mathcal{S}}), \quad \tilde{\mathcal{S}}=\{s\in \text{Memory}(\pi^*)\mid s\vert_{x_{ego}}<x_{int}\},$$
where $s\vert_{x_{ego}}<x_{int}$ represents all states in which the ego has not yet crossed the intersection. In practice we only have access to observations of states in the memory buffer hence we map the sampled observations, which contain the ego's position, back onto environment states which would result in the sampled observation. Even though we're not assured to map back onto the exact same environment state as was encountered in the optimal agent's state-space trajectory, this nevertheless increases the contingency agent's ability to be robust to initial states sampled from the optimal agent's trajectory. The optimal agent $\pi^*$ uses the standard environment initial state distribution: $p^{\pi^*}(s_0)=p(s_0)$.

In our experiments we found that a ratio of 50\% between regular and optimal-replay-buffer episode initializations gave the best results. lower values tended to decrease the contingency agent's ability to function well at the `hand-off' points, whereas higher values tended to overly impact the convergence of the contingency agent's parameters; $p(\tilde{s}_{\pi^*})$ adds a lot of variance in the state-space encountered by the contingency agent, and hence increases the complexity of correctly learning $Q^{\pi_1}$ over this larger domain. Thus we use the value $\beta=0.5$:
$$p^{\pi_1}(s_0)=\frac{1}{2}p(s_0) + \frac{1}{2}p(\tilde{s}_{\pi^*})$$ 

\subsection{Hierarchical Controller Algorithm}
\label{app:alg2}
Algorithm \ref{alg:h_controller} shows how the hierarchical controller chooses between available policies. The idea is for it to choose the policy with the highest estimated safety in the environment, using estimates of the environment dynamics.

\begin{algorithm}
\caption{Executing Hierarchical Controller}
\label{alg:h_controller}
\begin{algorithmic}[1]
\State $\Pi=\{\pi^*, \pi_1\}$ \Comment{available policies}
\State Init $o_0$ \Comment{initial env observation}
\State $\tau=\{o_0\}$
\While{episode not terminated}
    \For{$\pi\in \Pi$}
        \For{$i$ in simulation budget $M$}
            \State $\boldsymbol{\mu}_i\sim p(\boldsymbol{\mu}|\tau)$ \Comment{sample dynamics given observation history}
            \State $C^\pi_{\boldsymbol{\mu}_i}= \text{{\fontfamily{qcr}\selectfont Simulate}}(\pi, \boldsymbol{\mu}_i)$ \Comment{simulate env using $\boldsymbol{\mu}_i$, and env model}
        \EndFor
    \EndFor
    \State $\pi_{chosen}=\text{arg}\min_{\pi\in\Pi} \frac{1}{M}\sum_{i=1}^M C^\pi_{\boldsymbol{\mu}_i}$ \Comment{choose estimated safest policy}
    \State $a_t=\pi_{chosen}(o_t)$
    \State $o_{t+1}\sim \text{{\fontfamily{qcr}\selectfont Environment}}(a_t)$  \Comment{policy acts, env returns new observation}
    \State $\tau=\tau\cup \{o_{t+1}\}$
    \State Update $p(\boldsymbol{\mu}|\tau)$  \Comment{Update conditional probability on $\boldsymbol{\mu}$}
\EndWhile

\end{algorithmic}
\end{algorithm}

\subsection{Sampling Environment Dynamics}
\label{app:sampling_mu}
The quality of our estimation of collision probability depends on the quality of estimation of environment parameters. The estimation of the possible parameters is updated using the history of observations from the agent: $p(\boldsymbol{\mu}|\tau)$. We start with a uniform density on $\boldsymbol{\mu}$ at the start of each episode, and with every new observation $o_{t+1}$, we compare the actions taken by target vehicles, with the actions according to either possible target behaviour model ($b_i\in\{0,1\}$). If the observed target speed $v^{obs}$ is within a certain threshold $\varepsilon_v$ of the simulated behaviour speeds $v^{sim}_{b_i}$, then that target behaviour is retained in $p(\boldsymbol{\mu}|\tau)$. Using a stricter notation we can write $\boldsymbol{\mu}\vert_i=b_i$, where: 

$$p(b_i)=\mathcal{U}(B), \quad B=\{b\mid \left\vert v^{sim}_b-v^{obs}\right\vert< \varepsilon_v, b\in\{0, 1\}\},$$

where $v^{sim}_0, v^{sim}_1$ are the simulated speeds for a cooperative and aggressive target, respectfully. All $b_i$'s are independent, meaning there is no correlation between target behaviours.

To obtain our success rate and average score results, we sample without replacement $M=200$ values for $\boldsymbol{\mu}$, and average out both the success rate (i.e. how many times the ego successfully navigated the intersection without crashing), and the score (i.e. regular reward function in the environment).

\end{document}